\documentclass[10pt,twocolumn,letterpaper]{article}

\usepackage[accsupp]{axessibility}  % Improves PDF readability for those with disabilities.

\usepackage{iccv}              % To produce the CAMERA-READY version
\usepackage{algorithm}
\usepackage{algpseudocode} % 임시로 넣었는데 원래 쓰던 알고리즘 패키지로 수정
\usepackage{adjustbox}
\usepackage{booktabs}

\usepackage{amsmath, amssymb}
\newcommand{\paragrapht}[1]{\noindent\textbf{#1}}

\usepackage{soul}
\usepackage{tikz}
\usepackage{pifont}
\newcommand*\circled[1]{\tikz[baseline=(char.base)]{
            \node[shape=circle,draw,inner sep=0.5pt] (char) {#1};}}
\usepackage{xcolor}
\usepackage{array}

\definecolor{iccvblue}{rgb}{0.21,0.49,0.74}
\usepackage[pagebackref,breaklinks,colorlinks,allcolors=iccvblue]{hyperref}

\def\paperID{8782} % *** Enter the Paper ID here
\def\confName{ICCV}
\def\confYear{2025}

\title{OCK: Unsupervised Dynamic Video Prediction with Object-Centric Kinematics}
\author{
    Yeon-Ji Song\textsuperscript{1,2},  
    Jaein Kim\textsuperscript{1,2\thanks{Equal contribution} }, 
    Suhyung Choi\textsuperscript{1,2\footnotemark[1]}, 
    Jin-Hwa Kim\textsuperscript{2,3\thanks{Corresponding authors} },   
    Byoung-Tak Zhang\textsuperscript{1,2\footnotemark[2] }\\
    \textsuperscript{1}Seoul National University \quad
    \textsuperscript{2}SNU AIIS \quad
    \textsuperscript{3}NAVER AI Lab \\
    {\tt\small \{yjsong, jykim, s.choi\}@snu.ac.kr, j1nhwa.kim@navercorp.com, btzhang@snu.ac.kr}
}
\begin{document}
\maketitle
\begin{abstract}
Human perception involves decomposing complex multi-object scenes into time-static object appearance (\ie, size, shape, color) and time-varying object motion (\ie, position, velocity, acceleration).
% For machines to interact in the real world in a manner akin to human intelligence,
For machines to achieve human-like intelligence in real-world interactions, understanding these physical properties of objects is essential, forming the foundation for dynamic video prediction.
% and serves as a foundational rationale for dynamic video prediction.
% Recent advancements in object-centric transformers have shown promise for video prediction; however, they prominently focus on object appearance while neglecting motion, which is crucial for modeling dynamic interactions and ensuring temporal consistency in complex scenarios. 
While recent advancements in object-centric transformers have demonstrated potential in video prediction, they primarily focus on object appearance, often overlooking motion dynamics, which is crucial for modeling dynamic interactions and maintaining temporal consistency in complex environments.
To address these limitations, we propose OCK, a dynamic video prediction model leveraging object-centric kinematics and object slots.
We introduce a novel component named Object Kinematics that comprises explicit object motions, serving as an additional attribute beyond conventional appearance features to model dynamic scenes.
% We introduce a novel component named Object Kinematics that comprises low-level structured states of object motions, serving as an additional attribute beyond conventional appearance features.
The Object Kinematics are integrated into various OCK mechanisms, enabling spatiotemporal prediction of complex object interactions over long video sequences. Our model demonstrates superior performance in handling complex scenes with intricate object attributes and motions, highlighting its potential for applicability in vision-related dynamics learning tasks.
\end{abstract}

\iffalse
Human perception involves decomposing complex multi-object scenes into time-static object appearance (i.e., size, shape, color) and time-varying object motion (i.e., position, velocity, acceleration). For machines to achieve human-like intelligence in real-world interactions, understanding these physical properties of objects is essential, forming the foundation for dynamic video prediction. While recent advancements in object-centric transformers have demonstrated potential in video prediction, they primarily focus on object appearance, often overlooking motion dynamics, which is crucial for modeling dynamic interactions and maintaining temporal consistency in complex environments. To address these limitations, we propose OCK, a dynamic video prediction model leveraging object-centric kinematics and object slots. We introduce a novel component named Object Kinematics that comprises explicit object motions, serving as an additional attribute beyond conventional appearance features to model dynamic scenes. The Object Kinematics are integrated into various OCK mechanisms, enabling spatiotemporal prediction of complex object interactions over long video sequences. Our model demonstrates superior performance in handling complex scenes with intricate object attributes and motions, highlighting its potential for applicability in vision-related dynamics learning tasks.
\fi

% We introduce Kinematics Encoder to capture Object Kinematics, the motion states of objects over a structured image space.\
\section{Introduction} \label{intro}
\begin{figure}[t]
  \centering
  \includegraphics[width=1\linewidth]{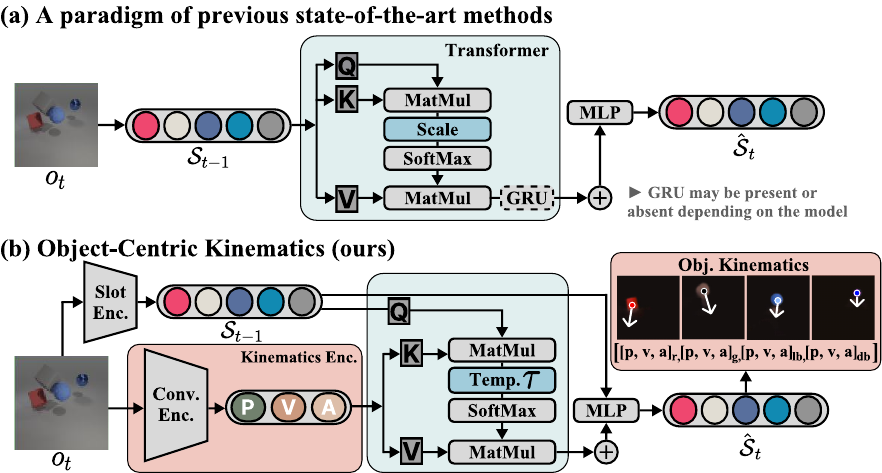}
  \caption{
  % Object-Centric Kinematics illustration.
  Comparison of OCK with previous slot-based video prediction methods utilizing transformers.
  The previous methods
  % , SlotFormer~\cite{wu2022slotformer} and OCVP~\cite{ocvp},
  leverage object attributes only.
  In contrast, OCK utilizes both object appearances and motions.
  }
  \label{fig:introduction}
\end{figure}

Human-level intelligence requires a deep understanding of the environment and the spatiotemporal interactions of surrounding objects~\cite{hamilton2022neuro}. This proficiency is fundamental to deep learning, as it underpins the human ability to perceive scenes as structured compositions of object components, facilitating object recognition in dynamically changing environments~\cite{kahneman1992reviewing}. % thereby recognizing objects in a dynamically changing world.
Despite advancements in modeling human intelligence, current approaches struggle to bridge the gap between human-level understanding and machine-generated world models, lacking the ability to accurately replicate human perception and interpretation of their surroundings.
To address this limitation, unsupervised object-centric representation learning has emerged as a promising framework for capturing the compositional structure of objects, mirroring the human ability to intuitively interpret their surroundings.
This approach has shown particular effectiveness in dynamic environments with diverse, constantly evolving, or previously unseen objects,
achieving remarkable performance across various domains, including scene understanding~\cite{phycine, steve}, object tracking~\cite{jiang2020generative,zhao2023ocmot}, reinforcement learning~\cite{smorl,srics} and video prediction~\cite{wu2022slotformer,gswm}.
% This approach has proven valuable in scenarios involving diverse, constantly changing, or previously unseen objects.
% similar to how humans instinctively understand their surroundings, demonstrating value in scenarios where the set of objectsencountered may be diverse and constantly changing, or even not previously encountered.
% It has provided
% Despite considerable focus on human intelligence, there remains a gap between human consciousness and the world model due to the limitations of machines in imitating how humans perceive the world and their surroundings.

Our work focuses on object-centric video prediction. The object-centric bias significantly improves the model's ability to capture object dynamics and interactions for successful video prediction, leading to enhanced predictive accuracy in complex environments. This approach optimally balances between expressivity and accuracy~\cite{ddlp}, enabling precise modeling of complex physical interactions (\ie, collision, freefall, mass and charge) that remain challenging for more general models, despite imposing constraints on the range of visual complexity it can effectively represent.
% offering improved predictive accuracy in targeted scenarios.
% It constrains the types of videos to be represented but enables accurate predictions of complex physical interactions such as collisions, that prove difficult for general models.
% While limiting in some aspects,   potentially leading to more accurate predictions in specific scenarios.
% 재인
% Especially, object-centric representation provides video prediction methods the inductive bias that discerns object dynamics and interactions in complex environments. Nevertheless, object centric approach should balance the trade-off between expressivity and accuracy~\cite{ddlp}, imposing constraints on the representable range of videos but precisely modeling challenging physical interactions such as collisions.

% 재인
% Recent object-centric approaches address this balance in video prediction by utilizing transformers, which have demonstrated efficient long-term prediction through attention mechanism and sequential positional encoding in diverse domains~\cite{vaswani2017attention}. Indeed, this functionality ~
Among various approaches to object-centric video prediction, transformers have demonstrated efficiency in addressing the trade-off between expressivity and long-term predictive accuracy through attention mechanisms and sequential positional encoding~\cite{vaswani2017attention}.
% Indeed,
This functionality of transformers combined with object-centric representation enables direct modeling of the spatiotemporal dynamics of objects within the object representation space~\cite{wu2022slotformer,ocvp}.
While prior works have achieved exceptional performance, certain approaches are restricted to simplistic environments, focusing solely on implicit object appearances~\cite{gswm}, while others can handle complex scenes but are often limited to short videos~\cite{ocvp, wu2022slotformer}.
See \cref{fig:introduction} for a visual explanation.
Successful prediction fundamentally relies on capturing \emph{long-term dependencies} and \emph{scene complexity}, including object interactions, dynamics, and visual variations.
The absence of either factor in prior works highlights the need for a model that effectively captures explicit physical interactions and temporal dependencies, with an emphasis on precise object motion modeling within video frames.
% by focusing on permutation equivariance among objects,

In this paper, we present Object-Centric Kinematics (OCK) for unsupervised object-centric video prediction.
We introduce Object Kinematics, which encapsulates object motion within a structured image space and plays a pivotal role in object tracking, demonstrating efficacy across both synthetic and real-world environments.
% a low-level structured state space of objects, which plays a central role in object tracking in complex environments.
The Object Kinematics are derived through two approaches:
The \textit{analytical approach} anticipates subsequent states from the input frame via logical reasoning and utilizes them for subsequent frame generation;
The \textit{empirical approach} leverages the given object information via inductive reasoning, focusing on its implicit learning.
We employ two versions of OCK to integrate the Object Kinematics with object appearances, named slots~\cite{locatello2020object}.
% , a set of object-centric representations extracted from the Slot Attention framework
We compare our model with the prior works that utilize object appearance information only~\cite{gswm,wu2022slotformer} in complex environments where object motion, appearance, and backgrounds vary repeatedly. 

In summary, we make the following contributions:
(1) We introduce an object-centric video prediction model leveraging Object Kinematics to comprehend time-varying object motion and time-static object appearance.
(2) We conduct comprehensive experiments with four recent slot-based methods on six synthetic and one real-world datasets. Our model significantly enhances the accuracy of predicting object dynamics in highly complex scenarios and exhibits better generalization to long sequences during test time.
(3) We investigate the impact and necessity of Object Kinematics through various ablation studies.
% (\ie, position, velocity, acceleration) (\ie, size, color, texture).
% can achieve state-of-the-art results on both
\section{Related work}
\subsection{Object-centric representation learning}
Recent unsupervised object-centric learning can be split into three approaches.
Spatial attention approaches employ CNN or spatial transformer networks to crop rectangular regions from an image, enabling the extraction of object attributes such as position, scale, or latent~\cite{iodine, chakravarthy2023spotlight,anderson2018bottom,kim2018bilinear,carion2020end}. However, these approaches often rely on a fixed-size sampling grid or coarse bounding box, which may not be suitable for scenes featuring diverse object sizes, potentially compromising training efficacy when the sampling grid fails to overlap with any object.
Sequential attention models, exemplified by RNN-based frameworks, sequentially attend to different regions in an image~\cite{burgess2019monet,engelcke2019genesis}, resulting in a suboptimal understanding of interrelationships between objects and image regions.
As a result, these models struggle to capture the global context of the image.
Lastly, iterative attention methods initialize a set of object representations, namely \textit{object slots}, and iteratively refine them to associate them with distinct regions of an image~\cite{slate,steve,wang2023slotvae}.
Predominantly inspired by Slot Attention~\cite{locatello2020object}, the iterative approach fosters competition among object slots by employing attention along the object dimension.
Our work focuses on the iterative attention approach, given its widespread usage for video prediction tasks.

\subsection{Dynamic video prediction}
Video prediction and generation in dynamic environments is a challenging task that has gained significant attention in recent years.
Approaches in video modeling include object-agnostic models, utilizing 3D convolutions~\cite{wang2018e3dlstm, gao2022simvp} or RNNs~\cite{oliu2018folded, wang2022predrnn}, and those employing structured or object-centric models, using probabilistic modeling techniques~\cite{phycine} or transformers~\cite{ning2023mimo, wu2022slotformer, ocvp,song2023learning}.
The object-agnostic models often require explicit human supervision, may be restricted to simple 2D datasets, and typically concentrate on modeling temporal changes using image features.
These approaches neglect explicit consideration of the composition of video frames or images.
In contrast, approaches utilizing object-centric representations to disentangle image frames into object attributes provide a more nuanced and comprehensive understanding beyond general representations.
Despite advancements in video prediction, generalization remains limited, with significant challenges in applying these models to unannotated data~\cite{wu2022optimizing}.
Inspired by video prediction models utilizing object-centric representations, our work diverges by investigating transformers' input components, architecture, and functioning.

\begin{figure*}[ht]
  \centering
  \includegraphics[width=0.95\linewidth]{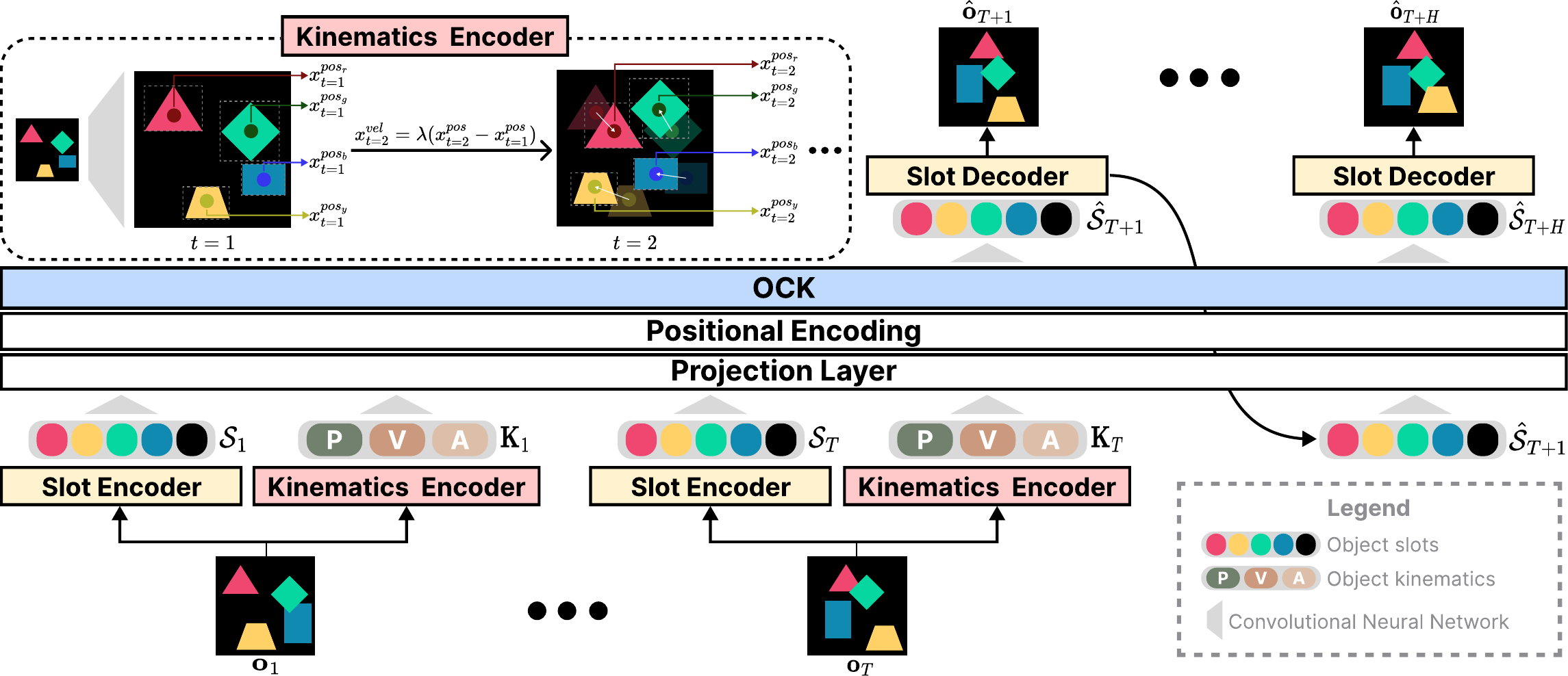}
  \caption{\textbf{Overview of our pipeline.} The model extracts object slots using the pretrained object-centric slot encoder, and Object Kinematics, including position, velocity, and acceleration, using the kinematics encoder, both from multiple video frames. These features are linearly projected and processed by the OCK mechanism with a positional encoding to generate future slots in an autoregressive manner.}
  \label{fig:architecture}
\end{figure*}

\subsection{Object-centric video prediction with transformers}
Our research builds on the foundations of SlotFormer~\cite{wu2022slotformer} and OCVP~\cite{ocvp}, both rooted in transformer architectures, to advance object-centric video prediction and generation. OCVP extends SlotFormer by slightly modifying the spatiotemporal attention block in two different ways.
The recurrent transformer strategy they utilize facilitates long-term predictions while mitigating the information loss on spatial object motion~\cite{selva2023video}. Additionally, their use of object-centric representations introduces an inductive bias for understanding object attributes within a sequence~\cite{ming2024surveyvp}.
However, their object-wise representations are biased toward invariant visual cues, focusing narrowly on implicit changes in video frames. This limited scope results in a lack of depth in capturing advanced object motions.
Extensive research has explored the generalization capabilities of transformers, particularly in symbolic mathematical modeling and compositional reasoning~\cite{mohnhaupt1991understanding,welleck2022symbolic,dziri2305faith}.
Building upon this area, our paper proposes a novel framework that comprehends spatiotemporal object dynamics and predicts the explicit sequence of time-varying object motions and time-static object appearances across various time steps.
\section{Method}
OCK is built upon an autoregressive object-centric transformer as summarized in \cref{fig:architecture}.
The input to the OCK is obtained through two modules, involving the extraction of object slots and Object Kinematics.
We demonstrate how the two features are subsequently fed into the OCK transformers, forming the basis for spatiotemporal future frame prediction and generation in an autoregressive way.

\subsection{Slot encoder} \label{sec:ocr}
We leverage the Slot Attention~\cite{locatello2020object} framework to map image features to permutation-invariant object slots.
% (\textit{ref.} Section~\ref{sec:object-centric_repr}).
A convolutional neural network extracts a grid of image features from a video frame $\mathbf{o}_t$, which are then flattened into a set of feature map $\mathbf{h}_t \in \mathbb{R}^{M \times D_\text{ft}}$ that represents its semantic information, where $M$ is the size of the flattened $H \times W$ feature grid and $D_\text{ft}$ is the dimensionality of the extracted feature maps. At each iteration, the model randomly initializes $N$ object slots $\mathcal{S}_0 \in \mathbb{R}^{N \times D_\text{slot}}$ and performs Slot Attention to update the object slots via iterative Scaled Dot-Product Attention~\cite{vaswani2017attention}.
This attention mechanism computes \text{softmax} over object slots and updates the slots with the weighted average to represent a part of the input. 
The output $\mathcal{S}_t$ is a set of object slots, which is fed into the autoregressive transformer along with the Object Kinematics.

\subsection{Kinematics encoder} \label{sec:kin}
We simultaneously perform Object Kinematics extraction, which involves a convolutional neural network (CNN) and numerical calculations to estimate the object motions.
The intuition behind this is to preserve the explicit properties of objects, thereby preventing the model from learning incorrect motions in the initial stages of training when scaling up to complex environments, which is a phenomenon often observed when relying solely on slots~\cite{wu2022slotformer,gswm}. 

% We extract the structured state embedding of each object that encompasses \emph{position}, \emph{velocity}, and \emph{acceleration}.
Initially, video frame $\mathbf{o}_t$ is processed by a CNN-based slot network $\phi$ to extract a grid of low-level image features, which are then utilized to generate low-dimensional object features $\mathbf{x}_t$ by localizing the center of mass of each object as 2D coordinates.
% Initially, we transform video frame $\mathbf{o}_t$ into low dimensional object features $\mathbf{x}_t$ via a CNN-based network $\phi$. This network extracts a grid of low-level image features, which are subsequently used to localize the center of mass of each object as 2D coordinates.
% of each object and then identifies the center of mass of each object as 2D coordinates.
The object features are passed through an MLP to ensure consistent encodings, forming the \textit{position} state denoted as $\mathbf{x}_t^\text{pos}$.
Note that $\mathbf{x}_t^\text{pos}$ is equivariant to the permutation over objects in $\mathbf{o}_t$.
% Note that any operation performed on these sets remains equivariant to the permutation of their objects.
The \textit{velocity} state $\mathbf{x}_t^\text{vel}$ is derived by computing the difference between two consecutive position states, $\mathbf{x}_t^\text{pos}$ and $\mathbf{x}_{t-1}^\text{pos}$.The \textit{acceleration} state $\mathbf{x}_t^\text{acc}$ is calculated via the differences between $\mathbf{x}_t^\text{vel}$ and $\mathbf{x}_{t-1}^\text{vel}$.
To ensure the consistency in the unified kinematic representation, a learnable parameter $\lambda$ is introduced to scale the velocity relative to the position, formulated as follows:
\begin{equation}
\mathbf{K}_t \triangleq
    \begin{bmatrix}
    \mathbf{x}^\text{pos}_t \\
    \mathbf{x}^\text{vel}_t \\
    \mathbf{x}^\text{acc}_t 
    \end{bmatrix} =
    \begin{bmatrix}
    \phi(\mathbf{o}_t) \\
    \lambda\big(\mathbf{x}^\text{pos}_t - \mathbf{x}^\text{pos}_{t-1}\big) \\
    \mathbf{x}^\text{vel}_t - \mathbf{x}^\text{vel}_{t-1}
    \end{bmatrix}.
    \label{eq:1}
\end{equation}

% The Object Kinematics $\mathbf{K}_t$ is constructed by linear concatenation of the three derived low-level geometric object features.
The Object Kinematics $\mathbf{K}_t \in \mathbb{R}^{N \times D_\text{kin}}$ is constructed by concatenating the low-level geometric object states and is subsequently fed as input to the transformer module along with the object slots $\mathcal{S}_t$. 
Notably, Object Kinematics is learned through the objects' centers extracted from the slot encoder, without reliance on task-specific loss functions, such as frame reconstruction or state transition objectives.
% Notably, Object Kinematics is learned in an unsupervised manner, without reliance on explicit object labels or task-specific loss functions, such as frame reconstruction or state transition objectives.
% $\mathbf{x}_t^\text{pos}$가 $\hat\mathbf{x}_t^\text{pos}$와 어떤 식으로 학습 되는지 간략히 기술
% independently of explicit object labels or loss functions, such as frame reconstruction or state transition.
Modeling kinematics in 2D image space is deliberate,
% as extending it to a 3D depth image using an additional network is computationally inefficient, due to the substantial computational costs involved in utilizing a pretrained optical flow model or frame reconstruction model during conversion
as extending it to a 3D depth image requires an additional network, leading to high computational costs due to the reliance on a pretrained optical flow or frame reconstruction model~\cite{karnewar2023holodiffusion,hu2023voxelflow}.
Furthermore, the predictive performance remains robust even in the absence of rotation and scaling factors.
Accordingly, the Object Kinematics are employed in two ways: $\textit{analytical}$ and $\textit{empirical}$ approach.

\paragrapht{Analytical approach}
leverages Object Kinematics from the video frame at time $t$ to predict the subsequent position state $\mathbf{x}^{\text{pos}'}_{t+1}$ based on the premise that objects will continue their motion patterns shortly.
OCK integrates current kinematics $\mathbf{x}_t^\text{pos}$ with the anticipated kinematics $\mathbf{x}^{\text{pos}'}_{t+1}$ to generate the subsequent frame $\hat{\mathbf{x}}_{t+1}^\text{pos}$.
% , as illustrated in Algorithm~\ref{algorithm}.
% This approach emphasizes continuous motion patterns based on logical reasoning by utilizing predicted kinematics information.
This approach leverages predicted kinematic information to emphasize continuous motion patterns through a reasoning-driven method.
% facilitating accurate predictions of future frames and efficient forecasting of Object Kinematics in complex and dynamic environments.

\begin{figure}[t]
  \centering
  \begin{subfigure}{0.32\linewidth}
    \centering
    \includegraphics[width=\linewidth]{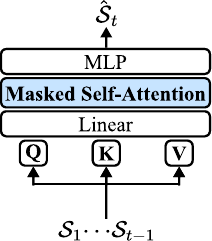}
    \caption{Transformer}
    \label{fig:trad}
  \end{subfigure}
  \hfill
  \begin{subfigure}{0.32\linewidth}
    \centering
    \includegraphics[width=\linewidth]{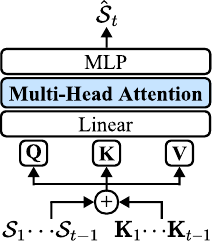}
    \caption{Joint-OCK}
    \label{fig:mha}
  \end{subfigure}
  \hfill
  \begin{subfigure}{0.32\linewidth}
    \centering
    \includegraphics[width=\linewidth]{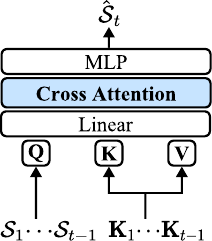}
    \caption{Cross-OCK}
    \label{fig:cca}
  \end{subfigure}
  \caption{An illustration of the OCK architecture. (a) Traditional transformer model processes slot exclusively. (b) Joint-OCK incorporates concatenated slots and kinematics as input. (c) Cross-OCK employs cross-attention, where slot representations serve as queries, and kinematics functions as keys and values.}
  \label{fig:vps}
\end{figure}

\paragrapht{Empirical approach}
relies solely on the current video frame $\mathbf{x}_t$ processed through an inductive gradient-based learning without additional dependencies.
% , relying on the learning process of transformers.
Specifically, it extracts Object Kinematics $\mathbf{K}_t$ from the video frame similarly, but focusing exclusively on the kinematics information at time $t$.
The motion pattern inferred from the current frame is utilized to predict the subsequent frame $\hat{\mathbf{x}}_{t+1}^\text{pos}$.
% , and utilizes it to forecast Object Kinematics for the subsequent time step.
% We train our model using both approaches with a slight emphasis on the empirical approach as it provides stable performance across model variations and dataset complexities.
% Our work focuses on the empirical approach as it provides stable performance across diverse video frame complexity.

\noindent\textbf{Remark}: The analytical approach is grounded in structured logical reasoning, while the empirical approach employs a gradient-based learning as illustrated in \cref{algorithm}.

\begin{algorithm}[t]
    \centering
    \caption{Object-Centric Kinematics (OCK)}
    \label{algorithm}
    \begin{algorithmic}[1]
    \State \textbf{Inputs:} input frames $\texttt{input}$, learnable queries $\texttt{init}$, number of iterations $T$
    \State \textbf{Require:} time difference $\delta$, learnable parameter $\lambda$
    \State \textbf{Modules:} OCK module $\texttt{OCK}(\cdot,\cdot)$, object slot module $\texttt{SA}(\cdot,\cdot)$, Object Kinematics module $\texttt{KIN}(\cdot)$
    \State $\texttt{slots} \leftarrow \texttt{init}$
    \For{$i \leftarrow 1$ to $T$}
        \State $\texttt{slots} \leftarrow \texttt{SA}(\texttt{slots}, \texttt{input})$
        \State $\texttt{kins} \leftarrow \texttt{KIN}(\texttt{input})$ \footnotemark
        % \\ \Comment{Note $\texttt{kins} = [\texttt{kins}_\text{pos},\texttt{kins}_\text{vel},\texttt{kins}_\text{acc}]$}
        \If{\text{Analytical Approach}}
            \State $\texttt{kins}^\prime_\text{pos} \leftarrow \texttt{kins}_\text{pos} + \texttt{kins}_\text{vel} \times \delta$
            \State $\texttt{kins}^\prime_\text{vel} \leftarrow \lambda (\texttt{kins}^\prime_\text{pos} - \texttt{kins}_\text{pos})$
            \State $\texttt{kins}^\prime_\text{acc} \leftarrow \texttt{kins}^\prime_\text{vel} - \texttt{kins}_\text{vel}$
            \State $\texttt{kins}^\prime \leftarrow [\texttt{kins}^\prime_\text{pos},\texttt{kins}^\prime_\text{vel},\texttt{kins}^\prime_\text{acc}]$
            \State $\texttt{slots}^\prime \leftarrow \texttt{OCK}(\texttt{slots}, [\texttt{kins}, \texttt{kins}^\prime])$
        \ElsIf{\text{Empirical Approach}}
            \State $\texttt{slots}^\prime \leftarrow \texttt{OCK}(\texttt{slots}, \texttt{kins})$
        \EndIf
        \State $\texttt{slots} \leftarrow \texttt{slots}^\prime$
    \EndFor
    \State \textbf{return} $\texttt{slots}^\prime$
    \end{algorithmic}
\end{algorithm}
% \footnotetext{Note that $\texttt{kins}$ is composed of position $\texttt{kins}_\text{pos}$, velocity $\texttt{kins}_\text{vel}$, and acceleration state $\texttt{kins}_\text{acc}$ of the current video frame.}
\footnotetext{Note that $\texttt{kins} = [\texttt{kins}_\text{pos}, \texttt{kins}_\text{vel}, \texttt{kins}_\text{acc}]$ of the current frame.}

\subsection{Autoregressive OCK transformers} \label{sec:trans}
OCK leverages the object slots $\mathcal{S}_t$ and the Object Kinematics $\mathbf{K}_t$ to predict object slots at the subsequent timestep $\hat{\mathcal{S}}_{t+1}$, while preserving temporal consistency.
% The predicted object slots $\hat{\mathcal{S}}_{t+1}$ are recursively fed back into the transformer, enabling an autoregressive prediction of future object slots $\hat{\mathcal{S}}_{t+h}$ over successive timesteps.
% This iterative process facilitates the model to dynamically refine its future predictions
In this work, we introduce two variants of the OCK mechanism, $\textit{Joint-OCK}$ and $\textit{Cross-OCK}$ as shown in \cref{fig:vps}.
% , each designed to enhance the modeling of spatiotemporal dependencies.
% The structural distinctions between these mechanisms are detailed in the supplementary material.
% Please refer to the supplementary material for details on the structural differences between the two transformer mechanisms.

\paragrapht{Joint-OCK.}
Similar to \citet{wu2022slotformer}, Joint-OCK adopts a standard transformer encoder. 
As shown in \cref{fig:mha}, the input formed by concatenating object slots and Object Kinematics undergoes a linear projection to align with the transformer's inner dimension $d_k$, followed by temporal positional encodings to preserve permutation equivariance, ensuring that object slots at the same timestep share the positional encodings for consistent object representations.
\begin{equation}
    \begin{aligned}
        \mathbf{q} = \mathbf{X}_t\mathbf{W}_q, \quad 
        \mathbf{k} &= \mathbf{X}_t\mathbf{W}_k, \quad 
        \mathbf{v} = \mathbf{X}_t\mathbf{W}_v, \\
        \mathrm{s.t.} \quad \mathbf{X}_t &= \big[\mathcal{S}_t,\, \mathbf{K}_t\big], \\
        \text{Joint-OCK}(\mathbf{v},\mathbf{k}, \mathbf{q}) &= \mathbf{v} \cdot
        \text{softmax} \left( \frac{\mathbf{k}^\top \mathbf{q}}{\sqrt{d_k}} \right).
    \end{aligned}
\label{eq:joint-ock}
\end{equation}

\paragrapht{Cross-OCK.}
Alternatively, Cross-OCK utilizes a cross-attention mechanism to integrate object slots $\mathcal{S}_t$ with the Object Kinematics $\mathbf{K}_t$ as illustrated in \cref{fig:cca}.
Inspired by transformer-based multi-scale feature learning~\cite{chen2021crossvit}, object slots with larger token sizes as $\texttt{queries}$ and Object Kinematics with smaller token sizes as $\texttt{keys}$ and $\texttt{values}$ achieve enhanced computational efficiency, thereby generating plausible future video frames.
% , facilitating effective information exchange between the two representations as follows:
% to improve computational efficiency thereby generating plausible future video frames.
% The \text{softmax} operation is applied along the last dimension, capturing the significance of each \texttt{key-value} pair for every \texttt{query}. The values are then aggregated via a weighted sum to generate the predicted frames. Unlike prior works that scale down weights by the dimension of $\texttt{keys}$, our framework modulates weights by a temperature parameter $\tau$, which modulates the inner products before the \text{softmax} operation ensuring precise calibration within the attention mechanism~\cite{ali2021xcit}.
The \text{softmax} operation is applied along the last dimension to capture the significance of each \texttt{key-value} pair for every \texttt{query}. The values are then aggregated via a weighted sum to generate the predicted frames. Unlike prior works that scale down weights by $\texttt{keys}$, our framework uses a temperature parameter $\tau$ to modulate inner products before \text{softmax}, ensuring precise attention calibration~\cite{ali2021xcit} as follows:
% Mathematically, the cross-attention \texttt{CA} mechanism is formulated 
\begin{equation}
\begin{aligned}
    \mathbf{q}=\mathcal{S}_t\textbf{W}_q, \quad
    \mathbf{k}=\mathbf{K}_t\textbf{W}_k, \quad
    \mathbf{v}=\mathbf{K}_t\textbf{W}_v, \\
    \text{Cross-OCK}(\mathbf{v},\mathbf{k}, \mathbf{q}\,;\,\tau)=\mathbf{v} \cdot
    \text{softmax}(\frac{\mathbf{k}^\top \cdot \mathbf{q}}{\tau}).
\end{aligned}
\label{eq:cross-ock}
\end{equation}

% The output of the transformers is the updated object slot representations according to their anticipated histories.
The following attention strategies allow the model to handle complex datasets effectively, ensuring robust performance. This proficiency enhances the model's ability to capture object interactions, discern intricate patterns, and generate comprehensive future frames. Please refer to the supplementary material for a detailed description.

\subsection{Model training} \label{sec:training}
OCK is trained in two steps utilizing a pretrained SAVi~\cite{savi} model.
Initially, we train SAVi to decompose video frames into object slots. Then, our model is trained by feeding the extracted object slots into OCK to predict future slots. The predicted slots are transformed into images and masks using the pretrained SAVi for frame reconstruction.
% , which are combined to generate the subsequent video sequence.

Our model undergoes training by taking the last $N$ output slots from the transformer. These features are then fed into a linear layer to output object slots at the subsequent timestep. To ensure the continuity of future frame prediction, following prior works~\cite{wu2022slotformer, ocvp}, the generated object slots $\hat{\mathcal{S}}_{T+1}$ serve as input for the subsequent slot prediction $\hat{\mathcal{S}}_{T+2}$
% along with the ground-truth object slots
in an autoregressive way, facilitating the generation of any number of future frames $H$.
The model objective is to minimize the object reconstruction loss and image reconstruction loss with a hyperparameter $\alpha$ as follows:
% which reduces the train-test discrepancy and enables longer predictions in complex scenarios.
% The model predicts all object slots at a single timestep in parallel; thereby, train using the predicted slots as inputs by minimizing both object reconstruction loss and image reconstruction loss, which reduces the train test discrepancy and enables longer prediction in complex scenarios.
\begin{equation}
    \mathcal{L} = \mathcal{L}_\text{object} + \alpha \mathcal{L}_\text{image}.
    \label{eq:training}
\end{equation}

% More precisely,
The object reconstruction loss $\mathcal{L}_\text{object}$ is computed as the L2 loss between the ground-truth object slots $\mathcal{S}_{T+h}$ and the reconstructed object slots $\hat{\mathcal{S}}_{T+h}$ as:
\begin{equation}
    \mathcal{L}_\text{object} = \frac{1}{N \cdot H}\sum_{n=1}^{N}\sum_{h=1}^{H}\| \hat{\mathcal{S}}^n_{T+h} - \mathcal{S}^n_{T+h} \|_2.
    % + \frac{1}{K}\sum_{k=1}^{K}\| f_{dec}(S_{T+k}) - x_{T+k} \|^2
    \label{eq:slotloss}
\end{equation}

The image reconstruction loss $\mathcal{L}_\text{image}$ is calculated using a frozen SAVi decoder $f_\theta^\text{SAVi}$ to transform the predicted object slots $\hat{\mathcal{S}}_{T+h}$ into an image, which is then compared to the corresponding ground-truth video frame $\mathbf{o}_{T+h}$ as:
\begin{equation}
    \mathcal{L}_\text{image} =\frac{1}{H}\sum_{h=1}^{H} \| f_\theta^\text{SAVi} (\hat{\mathcal{S}}_{T+h} ) - \mathbf{o}_{T+h} \|_2.
    % + \frac{1}{K}\sum_{k=1}^{K}\| f_{dec}(S_{T+k}) - x_{T+k} \|^2
    \label{eq:imgloss}
\end{equation}
\section{Experiments}
% First, we brieﬂy introduce the datasets and baselines. Then we evaluate the prediction quality and its video decomposition results. Additionally, we assess the efficacy of our model for long-term prediction. Lastly, we conduct ablation studies on OCK and the transformer components to explore the potential impact of architectural differences on overall performance.
% We first overview the datasets and baseline models.
This section evaluates OCK on synthetic and real-world datasets with diverse scene complexities.
We provide an overview of the datasets and baselines, followed by an evaluation of video prediction and scene decomposition accuracy.
Then, we conduct ablation studies to investigate the impact of architectural variations on overall performance.
% analyze our model's effectiveness in long-term predictions and

\begin{figure*}[ht]
  \begin{subfigure}{0.57\linewidth}
    \centering
    \includegraphics[width=1\linewidth]{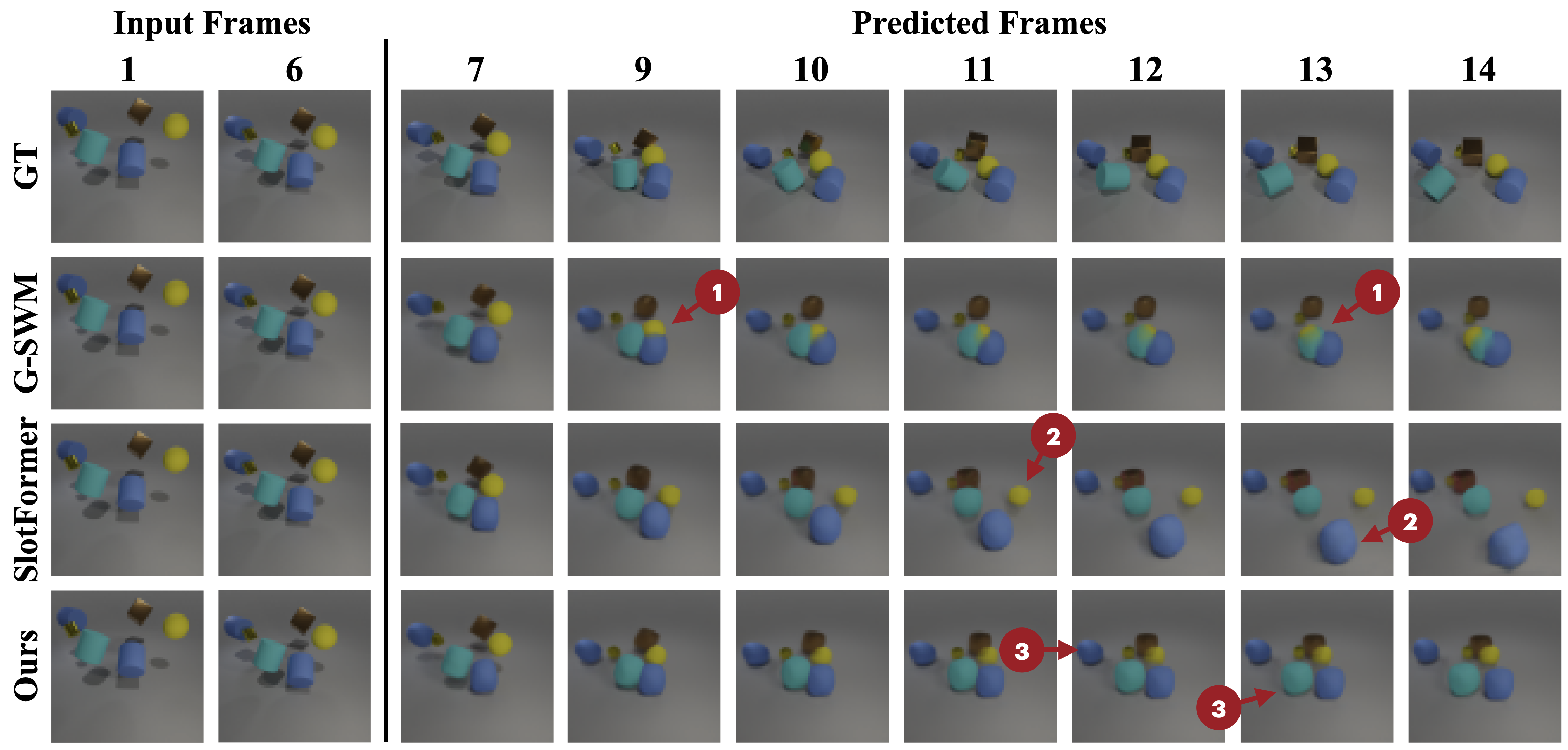}
    \caption{Generation results on MOVi-A. Frame 8 omitted due to its similarity to frame 7.}
  \label{fig:generation}
  \end{subfigure}
  \hfill
  \begin{subfigure}{0.42\linewidth}
    \centering
    \includegraphics[width=1\linewidth]{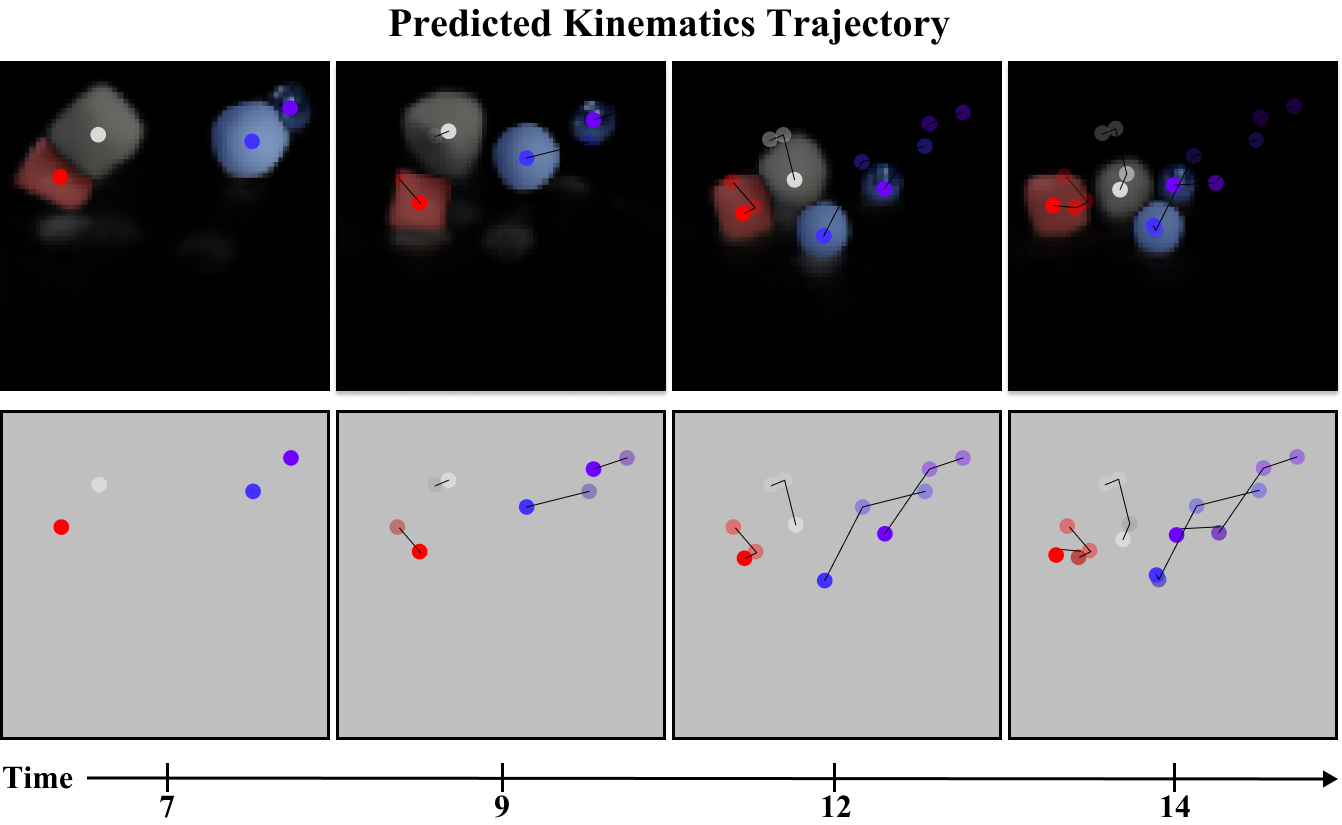}
    \caption{Predicted kinematics segment of Cross-OCK on MOVi-A.}
    \label{fig:kinematics}
  \end{subfigure}
  \caption{Generation results on MOVi-A.
  % We train the model with 6 frames and generate 8 frames, which is claimed sufﬁcient for the model to learn accurate dynamics.
  (a) We display the ground-truth frames (top row) and prediction results from two baseline models and Cross-OCK for comparison.
  (b) We visualize the kinematics trajectories of foreground objects across predicted frames. The top row displays the predicted kinematic trajectory overlaid on the frame, and the bottom row shows the isolated trajectory. Darker dots denote object positions in the latest predicted frame, while lighter dots, with progressively fading colors, represent earlier predictions.}
  \label{fig:fullgen}
\end{figure*}
% Please zoom in for the best view.
\begin{table*}[ht]
    \centering
    \setlength{\tabcolsep}{0.4mm}
    \resizebox{1\linewidth}{!}{%
    \begin{tabular}{@{}lcccccccccccccccccc@{}}
    \toprule[1.2pt]
    \multicolumn{1}{c}{} & 
    \multicolumn{3}{c}{OBJ3D} &
    \multicolumn{3}{c}{MOVi-A} &
    \multicolumn{3}{c}{MOVi-B} &
    \multicolumn{3}{c}{MOVi-C} &
    \multicolumn{3}{c}{MOVi-D} &
    \multicolumn{3}{c}{MOVi-E} \\
    \cmidrule(rl){2-4} \cmidrule(rl){5-7} \cmidrule(rl){8-10} \cmidrule(rl){11-13} \cmidrule(rl){14-16} \cmidrule(rl){17-19} 
    \textbf{Model} &
    {PSNR↑} & {SSIM↑} & {LPIPS↓} & {PSNR↑} & {SSIM↑} & {LPIPS↓} & {PSNR↑} & {SSIM↑} & {LPIPS↓} &
    {PSNR↑} & {SSIM↑} & {LPIPS↓} & {PSNR↑} & {SSIM↑} & {LPIPS↓} & {PSNR↑} & {SSIM↑} & {LPIPS↓}\\
    \midrule[1pt]                                                                   
    G-SWM       &31.142&0.900&0.039 & 26.140&0.784&0.133 & 21.850&0.677&0.247&19.466&0.451&0.554&20.567&0.548&0.355&21.166&0.534&0.359\\
    SlotFormer  &33.083&\underline{0.932}&\underline{0.024} & 25.180&0.785&0.134 &
                   21.329&0.690&0.215&19.482&0.456&0.534&20.675&\underline{0.565}&0.332&21.269&0.547&\underline{0.335}\\
    OCVP-Seq    &33.100&\underline{0.932}&0.025&26.240&0.789&0.127&\textbf{21.978}&\underline{0.701}&0.219&17.945&0.415&0.631&&Diverge&&&Diverge&\\
    OCVP-Par    &32.990&0.931&0.025&26.310&0.788&0.127&\underline{21.909}&0.688&0.226&17.941&0.402&0.650&&Diverge&&&Diverge&\\
    \midrule
    Joint-OCK  &\textbf{35.125}&\textbf{0.958}&\textbf{0.019} & \underline{27.259}&\underline{0.811}&\underline{0.124}& 
                 21.646&0.695&\textbf{0.198}&\underline{21.038}&\textbf{0.593}&\textbf{0.370}&
                \underline{22.087}&0.557&\underline{0.282}&\underline{22.394}&\underline{0.569}&\textbf{0.302}\\
    Cross-OCK &\underline{34.097}&0.925&\textbf{0.019}&\textbf{27.576}&\textbf{0.812}&\textbf{0.123}&
                {21.482}&\textbf{0.703}&\underline{0.209}&\textbf{21.040}&\underline{0.592}&\underline{0.376}&
                 \textbf{22.338}&\textbf{0.568}&\textbf{0.236}&\textbf{22.340}&\textbf{0.572}&\textbf{0.302} \\
    \bottomrule[1.2pt]
    \end{tabular}}
    \caption{Evaluation of video prediction quality across six synthetic datasets, increasing in scene complexity from left to right. ``Diverge" denotes the phenomenon where prediction performance degrades due to suboptimal slot extraction by the encoder.}
    \label{tab:video-prediction}
\end{table*}
\begin{table*}[ht!]
    \centering
    \setlength{\tabcolsep}{3.5mm}
    \resizebox{1\linewidth}{!}{%
    % \begin{adjustbox}{width=\textwidth, height=1.66cm, center}
    \begin{tabular}{lcccccccccc}
    \toprule[1.2pt]
    \multicolumn{1}{c}{} & \multicolumn{2}{c}{MOVi-A} & \multicolumn{2}{c}{MOVi-B} & \multicolumn{2}{c}{MOVi-C} &
    \multicolumn{2}{c}{MOVi-D} & \multicolumn{2}{c}{MOVi-E} \\
    \cmidrule(rl){2-3}\cmidrule(rl){4-5} \cmidrule(rl){6-7} \cmidrule(rl){8-9} \cmidrule(rl){10-11}
    \textbf{Model} & {FG-ARI↑} & {mIOU↑} & {FG-ARI↑} & {mIOU↑} & {FG-ARI↑} & {mIOU↑} & {FG-ARI↑} & {mIOU↑}& {FG-ARI↑} & {mIOU↑}\\
    \midrule[1pt]
    G-SWM      &0.431&0.500&0.410&0.443&0.238&0.414&0.239&\underline{0.327}&0.374&\underline{0.012}\\
    SlotFormer  &0.452&0.505&0.398&0.444&0.240&0.415&0.235&0.325&0.368&0.011\\
    OCVP-Seq &0.433&0.503&0.403&0.448&0.121&0.410&\multicolumn{2}{c}{Diverge}&\multicolumn{2}{c}{Diverge} \\
    OCVP-Par &0.435&0.501&0.364&0.449&0.113&0.402&\multicolumn{2}{c}{Diverge}&\multicolumn{2}{c}{Diverge} \\
    \midrule
    Joint-OCK   &\underline{0.560}&\underline{0.541}&\textbf{0.483}&\textbf{0.453}&\textbf{0.347}&\textbf{0.528}&
                      \underline{0.429}&\textbf{0.482}&\underline{0.379}&\textbf{0.019}\\
    Cross-OCK   &\textbf{0.563}&\textbf{0.547}&\underline{0.481}&\underline{0.452}&\underline{0.339}&\underline{0.515}&
                    \textbf{0.430}&\textbf{0.482}&\textbf{0.380}&\textbf{0.020}\\
    \bottomrule[1.2pt]
    \end{tabular}}
    % \end{adjustbox}
    \caption{Evaluation of unsupervised video decomposition across five MOVi datasets. We exclude OBJ3D due to the absence of segmentation masks, which are necessary for generating the decomposition of individual slots.}
    \label{object-dynamics}
\end{table*}

\subsection{Experimental details}
\subsubsection{Datasets} \label{sec:dataset}
We evaluate the performance of our proposed module across synthetic and real-world datasets to ensure a comprehensive assessment across a spectrum of complex scenarios.

\paragrapht{OBJ3D}~\cite{gswm} consists of 3D geometric objects on a gray background, set in motion to simulate dynamic environments in synthetic scenes. Each video consists of 100 frames at a 128×128 resolution, with randomly colored objects rolling toward dynamically positioned central objects.

\paragrapht{Multi-Object Video (MOVi)}~\cite{kubric} progressively increases in complexity from A to E, introducing a wider variety of objects' appearances, motions, and backgrounds. % MOVi-A to MOVi-E
MOVi datasets are specifically designed to test object discovery and tracking in dynamic environments.
From an object-centric learning perspective, MOVi-\{C,D,E\} are regarded as challenging testbeds, as they consist of real-life objects (\ie, shoes, toys) and backgrounds (\ie, sky, grass, marble floor) captured from large camera motions.

\paragrapht{Waymo Open Dataset}~\cite{waymo} consists of high-resolution video sequences recorded at a resolution of 1280×1920 using a multi-camera system installed on Waymo vehicles. It includes 798 training scenes and 202 validation scenes, each lasting 20 seconds and captured at 10 frames per second.
% To balance computational efficiency with data quality, the dataset is downsampled to 5 fps for both training and validation.

% \paragrapht{Youtube-VIS 2022} dataset, \textcolor{red}{todo}
% Among diverse categories of the Youtube-VIS 2022 (YTVIS22) dataset, we use the animals only. E.g., ape, parrot, snake, duck, dog, sheep, monkey.

\subsubsection{Baselines} \label{sec:models}
\paragrapht{G-SWM}~\cite{gswm} is one of the representative models in the domain of dynamics prediction from images, placing particular emphasis on object-centric representations.
% by breaking down scenes into foreground and background components.
% , while also disentangling object appearances. 
It models object interactions using appearance information through a graph neural network and employs hierarchical latent modeling to capture temporal dynamics over time.
% This unique capability allows it to model object interactions and dynamics.
% Through the adoption of object-centric representations, the model acquires a nuanced understanding of how individual objects influence the overall dynamics, making it a powerful tool for tasks such as video prediction and scene understanding.

\paragrapht{SlotFormer}~\cite{wu2022slotformer} (current \textbf{SOTA}) is designed for unsupervised visual dynamics simulation with object-centric representations. It leverages the Slot Attention framework to extract a set of slots that encode object attributes and their spatial relationships within a video frame.
% representing various components of a visual scene, which encapsulate object attributes and their spatial relationships.

\paragrapht{OCVP}~\cite{ocvp} is an extension of SlotFormer that further explores the transformers by separating the attention block into specialized temporal and relational attention blocks. OCVP-Seq refers to the sequential processing of temporal and relational attention blocks, while OCVP-Par refers to the parallel processing of the two attention blocks.

\subsubsection{Evaluation metrics} \label{sec:metrics}
\paragrapht{Video Prediction.} To assess the visual quality of the predicted videos, we report PSNR~\cite{huynh2008psnr}, SSIM~\cite{wang2004images}, and LPIPS~\cite{lpips}, where LPIPS demonstrates the highest perceptual similarity to human perception compared to PSNR and SSIM. Despite the limitations of PSNR and SSIM in accurately assessing video quality~\cite{sara2019image}, our model demonstrates strong performance in all metrics.

\paragrapht{Scene Decomposition.} We calculate the foreground Adjusted Rand Index (FG-ARI) and mean Intersection over Union (mIoU) to evaluate the predicted object dynamics using per-slot object masks generated by the SAVi decoder. FG-ARI is a similarity metric that assesses the correspondence between predicted and ground-truth segmentation masks in a permutation-invariant manner, renowned for its comprehensive evaluation of the predicted dynamics~\cite{savi}.

\begin{figure}[t]
    \centering
    \includegraphics[width=1\linewidth]{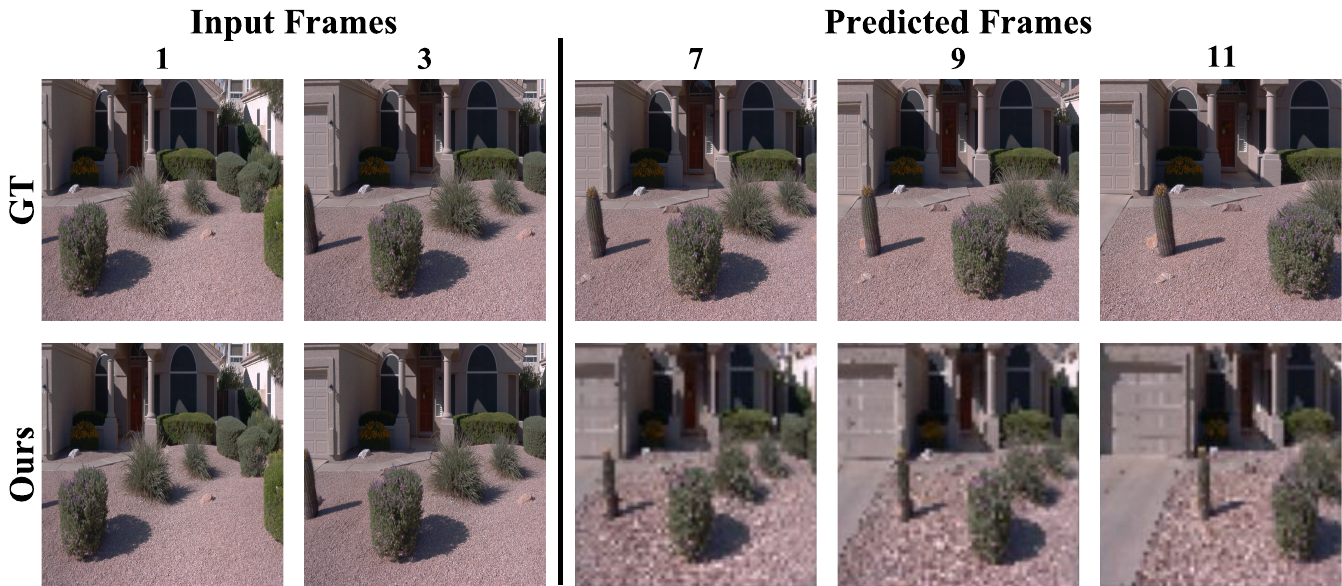}
    \caption{Generation results of OCK on Waymo Open Dataset.
    % The predicted frames demonstrate the successful extraction of objects to generate plausible future frames.
    }
    \label{fig:generation_waymo}
\end{figure}
\begin{table}[t]
  \centering
  \resizebox{0.72\linewidth}{!}{%
  \begin{tabular}{lccc}
        \toprule[1.2pt]
        \textbf{Model} & {PSNR↑} & {SSIM↑} & {LPIPS↓} \\
        \midrule[1pt]
        % G-SWM & & & \\
        SlotFormer & 19.127 & 0.330   & 0.714 \\
        OCVP-Seq   & 18.983 & 0.329   & 0.718 \\
        OCVP-Par   &        & Diverge &      \\
        \midrule
        Joint-OCK  & \underline{25.023} & \textbf{0.798} & \underline{0.251}  \\
        Cross-OCK  & \textbf{25.979} & \underline{0.728} & \textbf{0.220} \\
        \bottomrule[1.2pt]
    \end{tabular}
  }
  \caption{Evaluation of video prediction on Waymo Open dataset against transformer-based baseline models. }
  \label{tab:waymo_results}
\end{table}

\subsection{Video prediction on synthetic dataset} \label{sec:video}
% \Cref{tab:quality} presents the quantitative results of the predicted video frames across six datasets.
OCK demonstrates robust performance across environments of varying complexity,
showcasing its ability to generate perceptually realistic predictions closely aligned with human perception as presented in \cref{tab:video-prediction}.
Notably, Cross-OCK achieves superior performance on complex MOVi-\{D,E\} datasets by enabling the model to focus on relevant features specific to individual objects.
% , as this mechanism enables the model to target relevant features for specific objects.
We present qualitative results in \cref{fig:generation}, where our model successfully generates future frames despite minor discrepancies (marked \circled{3}) in certain object locations compared to the ground-truth frames.
These subtle differences have minimal impact on the overall prediction quality.
In contrast, baseline models exhibit significant shortcomings, such as objects collapsing into amorphous shapes (marked \circled{1}) or losing their appearance and generating incorrect dynamics (marked \circled{2}).
\cref{fig:kinematics} showcases the model's ability to accurately extract kinematics features from a sequence of video frames, facilitating precise prediction of object interaction. 
% where several objects collapse into a formless shape, denoted \circled{1}, or where an object loses its appearance and creates wrong dynamics, denoted as \circled{2}.

Our model consistently demonstrates the most reliable and visually plausible predictions. Both OCK models strategically prioritize key features by integrating high-level object embeddings with low-level visual cues, facilitating precise object slot extraction and underscoring the pivotal role of Object Kinematics in preserving coherent appearances and motions for accurate video prediction.
% This targeted strategy ensures prioritization of the most pertinent features, rather than evenly spreading attention, which assists in extracting object slots by linking high-level object embeddings with low-level visual cues.
% This result highlights the importance of Object Kinematics and the ability of our model to maintain coherent object appearances and motions for video prediction.
% , thereby generating accurate future frames.
% as it demonstrates OCK's ability to maintain coherent object appearances and motions, thereby enhancing the accuracy of generated frames.

\subsection{Scene decomposition on synthetic dataset} \label{sec:object}
To evaluate the accuracy of the predicted frames, we assess the quality of per-slot video decomposition using segmentation masks obtained from the video prediction process.
As reported in \cref{object-dynamics}, OCK exhibits performance comparable to four baseline models and proves the validity of the visual quality of the predicted videos. Accordingly, this underscores the positive trend of our model surpassing well-established benchmarks. 
For qualitative per-slot decomposition results, please refer to the supplementary material.
% Please refer to the supplementary material for visualizations.
% \subsection{\textcolor{red}{Generalization real-world}}

% OCK consistently outperforms baseline models, demonstrating its practical effectiveness in modeling object kinematics.
% The Waymo Open Dataset, while primarily focused on real-world driving scenes and traffic-related elements, serves as a highly challenging benchmark for object-centric learning due to its inherent visual complexity.
% As a result, videos capturing high-speed vehicle motion and rapid background changes may affect performance, which is an unavoidable challenge in real-world driving conditions. Despite overall performance degradations, OCK demonstrates comparable results to baselines and validates the visual quality of the predicted videos.
\subsection{Evaluation on real-world dataset}
We evaluate video prediction performance in the Waymo Open Dataset in \cref{tab:waymo_results}.
We exclude scene decomposition due to the low reliability of the preprocessed segmentation masks. The Waymo Open Dataset, renowned for its visual complexity in real-world driving scenarios, serves as a challenging benchmark for object-centric learning.
Consequently, high-speed vehicle motion and rapid background changes introduce inherent challenges, potentially affecting performance in real-world driving conditions.
Despite these limitations, both OCK approaches achieve outstanding results compared to baseline models while preserving the visual quality of predicted videos.
Moreover, as shown in \cref{fig:generation_waymo}, OCK successfully captures object interactions and makes structured predictions of object attributes and dynamics within the scene.
For additional qualitative results, please refer to the supplementary material.

\subsection{Evaluation on long-term generalization} \label{sec:longer}
\begin{figure}[t]
    \centering
    \includegraphics[width=1\linewidth]{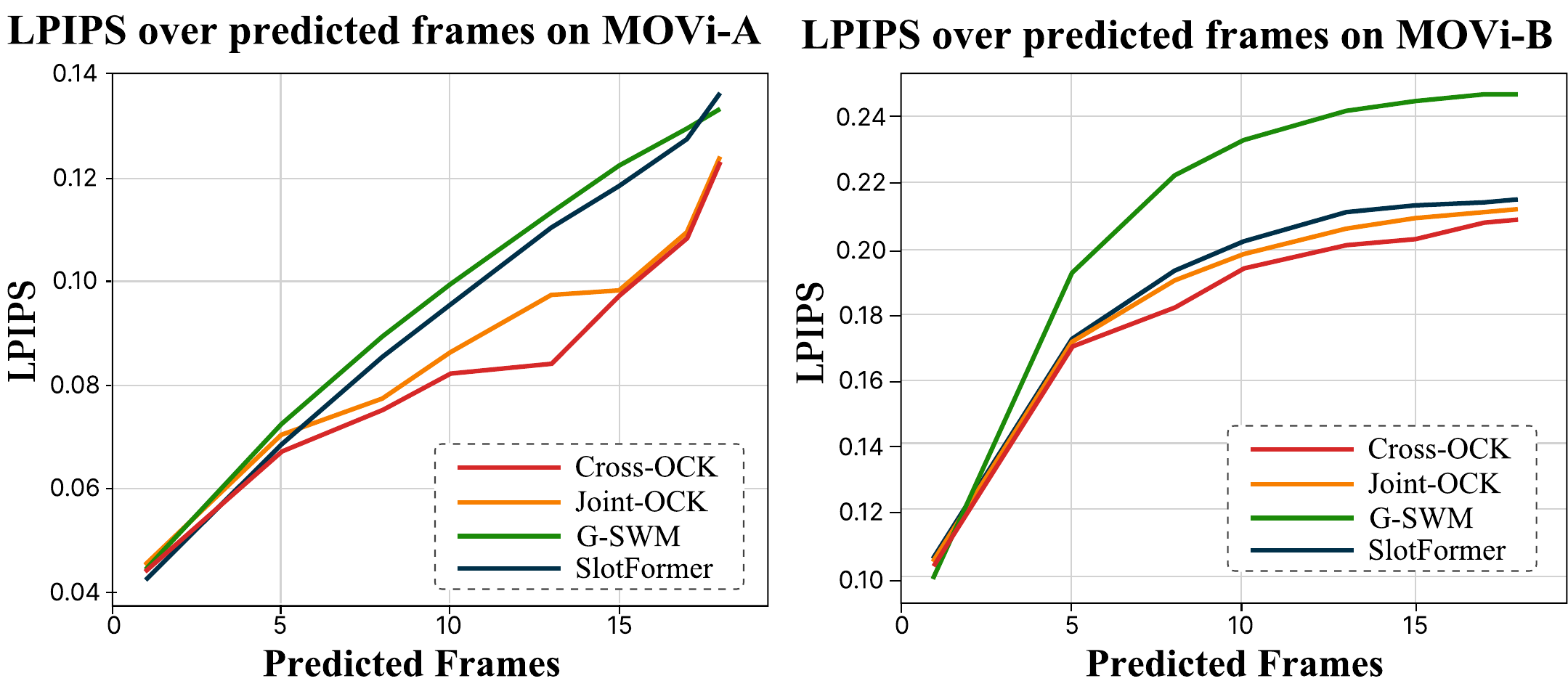}
    \caption{Long-term generalization results on the MOVi datasets, trained with 6 input frames and 8 future frames. Each video contains 24 frames, with generalization performance evaluated up to the 18th frame using the first 6 frames used as input.}
    \label{fig:longer}
\end{figure}

To test long-term prediction capabilities, we evaluate OCK on two MOVi datasets over a longer horizon, surpassing the settings used in training.
This is noteworthy as both datasets are trained with only six frames, significantly shorter than the full lengths.
Even without additional regularization, OCK exhibits strong generalization for longer sequences during test time.
In \cref{fig:longer}, the perceptual similarity of all models remains stable during initial frames but begins to diverge from the 7th predicted frame onward.
% OCK exhibits enhanced long-term predictive performance, surpassing baseline models in accuracy.
% OCK demonstrates superior long-term prediction accuracy, outperforming baseline models.
Unlike baseline models that rely solely on object appearances and are prone to error accumulation, OCK effectively generalizes to variations in object motion and appearance, enhancing its long-term dynamic video prediction capability.
% As anticipated, baseline models, which rely solely on object appearances without explicit modeling of object properties, experience performance degradation over time due to error accumulation. In contrast, OCK effectively generalizes to variations in object motion and appearance, enhancing its capability for long-term dynamic video prediction.

\subsection{Ablation studies} \label{sec:ablation}
In this section, we analyze how each transformer components influence the performance of dynamic video prediction in \cref{tab:components}.
We further conduct ablation studies on the two approaches, Joint-OCK and Cross-OCK, in \cref{tab:kinematics}.

\subsubsection{Transformer component analysis}
\paragrapht{Input frame.}
Our default training configuration for MOVi datasets involves utilizing six input frames and predicting eight frames. Increasing the number of input frames improves model performance; however, a slight decline is observed when further extending it to eight frames. This phenomenon arises from the adequacy of a six-frame history length for accurately capturing the object dynamics.

\paragrapht{Transformer (Trans.) layer.}
We utilize four transformer layers as the default for MOVi-A.
Additional layers result in unstable model training, and further augmentation causes diverging loss metrics.
This underscores the crucial balance between the depth of the transformer model and training stability in addressing the complexities of object dynamics.

\paragrapht{Vanilla positional encoding (P.E.).}
Our work incorporates temporal positional encoding such that object slots at the same timestep receive identical positional encoding.
To evaluate its significance, we compare our approach with sinusoidal positional encoding, which disrupts permutation equivariance among object slots.
As shown in \cref{tab:components}, preserving permutation equivariance serves as a crucial prior, thus, maintaining the equivariance is essential to ensure accurate long-term dynamic video prediction.

\paragrapht{Teacher forcing.}
We replace our model to incorporate a teacher forcing strategy~\cite{radford2018improving} by feeding ground-truth object slots instead of predicted object slots during training.
Surprisingly, this leads to a significant decline in performance, particularly affecting performance over longer prediction horizons.
Thus, it highlights the importance of learning to handle its imperfect predictions for accurate dynamics modeling during training to optimize real-world predictions.

\begin{table}[t]
    \centering
    \setlength{\tabcolsep}{3.3mm}
    \resizebox{0.9\linewidth}{!}{%
    \begin{tabular}{lccc}
        \toprule[1.2pt]
        % \multicolumn{1}{c}{} & 
        % \multicolumn{3}{c}{MOVi-A} \\
        % \multicolumn{3}{c}{Pred Length 14} \\
        % \multicolumn{1}{c}{}  
        % \cmidrule(rl){2-4}
        Method & {PSNR↑} & {SSIM↑} & {LPIPS↓}  \\
        \midrule[1pt]
        Cross-OCK(A)  &\textbf{27.576}&\textbf{0.812}&\textbf{0.123}\\
        Joint-OCK(A)  &\underline{27.259}&\underline{0.811}&\underline{0.124}\\
        \quad\quad Input Frame = 4    &27.012&0.801&0.125\\
        \quad\quad Input Frame = 8    &27.122&0.806&0.125\\
        \quad\quad Trans. Layer = 6   &26.924&0.796&0.130\\
        \quad\quad Trans. Layer = 8   &26.503&0.784&0.133\\
        \quad\quad Vanilla P.E.       &23.600&0.591&0.205\\
        \quad\quad Teacher Forcing    &23.583&0.589&0.207\\
        \bottomrule[1.2pt]
    \end{tabular}}
    \caption{Ablation study of transformer components on MOVi-A.}
    \label{tab:components}
\end{table}

\subsubsection{Object kinematics analysis}
We evaluate both analytical and empirical approaches to investigate the potential impact of Object Kinematics computation variations on overall performance in \cref{tab:kinematics}.
We train our model using six frames to predict eight subsequent frames and report the results for predicting 10 frames.
Our analysis reveals a slight performance advantage for the analytical approach over the empirical approach.
This is attributed to the fact that using the temporal positional state of the current frame for guidance can be advantageous, but object dynamics in complex environments are often aperiodic, such that an inductive gradient-based learning may introduce confusion.
Therefore, the analytical approach, which calculates the anticipated Object Kinematics of the subsequent video frame and integrates it with the current frame's kinematics, generates more accurate frames.
This is especially significant in complex dynamic environments where objects collide and consistently change their appearance since it enables the model to predict accurate dynamics and generate future frames with even greater precision.

\begin{table}[t]
  \centering
  \setlength{\tabcolsep}{0.5mm}
  \resizebox{1\linewidth}{!}{%
  \begin{tabular}{lcccccc}
        \toprule[1.2pt]
        \multicolumn{1}{c}{} & 
        \multicolumn{3}{c}{MOVi-A} & 
        \multicolumn{3}{c}{MOVi-B} \\ 
        \cmidrule(rl){2-4} \cmidrule(rl){5-7}
         Method & {PSNR↑} & {SSIM↑} & {LPIPS↓} & {PSNR↑} & {SSIM↑} & {LPIPS↓} \\
         \midrule[1pt]
         Cross-OCK(A)  &\textbf{27.576}&\textbf{0.812}&\textbf{0.123}&\underline{21.482}&\textbf{0.703}&\textbf{0.209}\\
         Joint-OCK(A)  &27.259&\underline{0.811}&\underline{0.124}&\textbf{21.494}&\underline{0.695}&\underline{0.212} \\
         \midrule
         Cross-OCK(E)  &\underline{27.536}&0.791&0.125&21.480&0.693&0.213  \\
         Joint-OCK(E)  &26.750&0.801&0.125&\underline{21.482}&0.694&0.213 \\
         \midrule
         Transformer   &25.180&0.785&0.134&21.329&0.690&0.215  \\
         \bottomrule[1.2pt]
    \end{tabular}
  }
  \caption{Kinematic ablation of both transformer mechanisms on the MOVi-\{A,B\} datasets. (A) for the analytical approach, and (E) denotes the empirical approach.}
  \label{tab:kinematics}
\end{table}
\section{Limitations}
% Successfully implementing our framework in real-world dynamic scenarios requires overcoming significant challenges.
Despite considerable advancement in realism compared to previous work, we find evidence of a significant gap in the current slot-based object-centric encoder, which hinders its scalability in in-the-wild datasets \cite{seitzer2022bridging}.
Our framework can easily integrate state-of-the-art pretrained slot encoders, thus, a possible solution is to substitute the slot encoder to extract object representations from real-life videos, which would be a significant contribution to future work.
% Scaling our approach to include such datasets holds significant potential for various downstream tasks.
% including robotics manipulation and action planning.

\section{Conclusion}
In this paper, we present OCK, an object-centric video prediction model that captures intricate object details by leveraging time-static object slots and time-varying Object Kinematics in an unsupervised manner.
We propose a novel component named Object Kinematics along with two transformer architectures to enhance dynamic video prediction performance in complex synthetic and real-world datasets.
% Experimental results demonstrate that our model effectively captures spatiotemporal object patterns by leveraging time-varying kinematic attributes, in contrast to prior approaches focused on static object appearances and explicit scene-level interactions.
Experimental results demonstrate that our model effectively captures spatiotemporal object patterns by utilizing time-varying kinematic attributes, surpassing prior methods that primarily emphasize static object appearances and explicit scene-level interactions.
Incorporating time-varying kinematic information shows significant potential for modeling object dynamics, enabling improved object-centric video prediction and generation in real-world environments.
% Our model demonstrates both efficacy and adaptability across a wide range of challenging scenarios, ensuring its relevance and applicability in real-world applications.

\section*{Acknowledgements}
This work was partly supported by the IITP (RS-2021-II212068-AIHub/10\%, RS-2021-II211343-GSAI/15\%, RS-2022-II220951-LBA/15\%, RS-2022-II220953-PICA/20\%), NRF (RS-2024-00353991-SPARC/20\%, RS-2023-00274280-HEI/10\%), and KEIT (RS-2024-00423940/10\%) grant funded by the Korean government.

{
    \small
    \bibliographystyle{ieeenat_fullname}
    \bibliography{main}
}

\end{document}